# Identifying Best Practice Melting Patterns in Induction Furnaces

## A Data-Driven Approach Using Time Series K-Means Clustering and Multi-Criteria Decision Making

Howard, Daniel Anthony; Jørgensen, Bo Nørregaard; Ma, Zheng Grace

*Published in:*
Energy Informatics

*DOI:*
10.1007/978-3-031-48649-4_16

*Publication date:*
2023

*Document version:*
Accepted manuscript

*Citation for pulished version (APA):*
Howard, D. A., Jørgensen, B. N., & Ma, Z. G. (2023). Identifying Best Practice Melting Patterns in Induction Furnaces: A Data-Driven Approach Using Time Series K-Means Clustering and Multi-Criteria Decision Making. In B. N. Jørgensen, L. C. Pereira da Silva, & Z. Ma (Eds.), *Energy Informatics: Third Energy Informatics Academy Conference, EI.A 2023, Campinas, Brazil, December 6–8, 2023, Proceedings, Part I* (pp. 271–288). Springer. https://doi.org/10.1007/978-3-031-48649-4_16

# Identifying Best Practice Melting Patterns in Induction Furnaces: A Data-Driven Approach Using Time Series K-Means Clustering and Multi-Criteria Decision Making

Daniel Anthony Howard[1[0000-0003-4556-0602]], Bo Nørregaard Jørgensen[1[0000-0001-5678-6602]], and Zheng Ma[1[0000-0002-9134-1032]]

[1] SDU Center for Energy Informatics, Maersk Mc-Kinney Moller Institute, University of Southern Denmark, Odense, Denmark
danho@mmmi.sdu.dk, zma@mmmi.sdu.dk, bnj@mmmi.sdu.dk

**Abstract.** Improving energy efficiency in industrial production processes is crucial for competitiveness, and compliance with climate policies. This paper introduces a data-driven approach to identify optimal melting patterns in induction furnaces. Through time-series K-means clustering the melting patterns could be classified into distinct clusters based on temperature profiles. Using the elbow method, 12 clusters were identified, representing the range of melting patterns. Performance parameters such as melting time, energy-specific performance, and carbon cost were established for each cluster, indicating furnace efficiency and environmental impact. Multiple criteria decision-making methods including Simple Additive Weighting, Multiplicative Exponential Weighting, Technique for Order of Preference by Similarity to Ideal Solution, modified TOPSIS, and VlseKriterijumska Optimizacija I Kompromisno Resenje were utilized to determine the best-practice cluster. The study successfully identified the cluster with the best performance. Implementing the best practice operation resulted in an 8.6% reduction in electricity costs, highlighting the potential energy savings in the foundry.

## 1 Introduction

The industrial sector accounts for approximately 40 % of global energy consumption and represents the second-largest contributor of $CO_2$ emissions following power generation [1]. The iron and steel industry is categorized as the industry sub-sector with the highest $CO_2$ emissions, accounting for approximately 2.6 GtCO2 in 2020 [2]. In November 2021, the International Energy Agency (IEA) reported that the industry sector is not on track to meet the Net Zero Emissions by 2050 scenario [2].

Several strategies and recommended actions have been proposed to accelerate the industry's progress toward meeting the Net Zero Emissions by 2050 scenario targets. The strategies include increased direct and indirect electrification of processes and improved overall energy efficiency through best-practice operation and maintenance [2]. Within the iron and steel sub-sector, IEA emphasizes the need for energy efficiency measures by deploying the best available technologies [3]. Furthermore, data collection and transparency should be emphasized to facilitate performance benchmarking assessments.

The foundry industry is a vital sector of the manufacturing industry that produces metal castings for a wide range of applications, including automobiles, infrastructure, and consumer goods. As a result of high demand, crude steel production doubled between 2000 and 2021 [4]. A report on global steel production costs showed that the production cost of one tonne of steel had increased by an estimated 51 % from 2019 to 2021. The report also identified that countries with low raw material and energy costs had lower production costs. As the foundry production process is highly energy-intensive and generates significant greenhouse gas emissions, there is an increasing need to improve energy efficiency and sustainability to increase the market competitiveness for European countries and meet the goals set forward by the IEA.

Denmark has proposed climate goals of reducing greenhouse gas emissions by 70 % by 2030 compared to 1990 levels and being climate neutral by 2050 [5]. The 15 Danish foundries produced 90 Mt of castings in 2019 across the industry, accounting for 1.5 % of Danish energy consumption and 3.2 % of Danish industrial energy consumption [6, 7]. Furthermore, the Danish government has agreed to phase in a $CO_2$ tax starting from 2025, which all industries must pay based on their emissions [8]. The foundries constitute a significant part of Danish energy consumption, and for Denmark to meet the climate goals, the foundry industry must become increasingly sustainable. Furthermore, electrical consumers have been shown to have an increasing alertness toward electricity price and $CO_2$ emissions [9].

The process energy consumption associated with the foundry process presented in Fig. 1 is mapped according to the approximate distribution identified by [10]. As seen in Fig. 1, the melting process accounts for approximately 55 % of the energy consumption in a casting process. Furthermore, the primary forming in the casting mold accounts for approximately 20 % of the energy consumption. It is, therefore, essential to address the melting and casting processes to improve a foundry process's energy efficiency and flexibility. The top drivers for energy efficiency covering the Swedish aluminum and casting industry were described by [11]. The drivers included the desire to reduce power charge, avoid exceeding power peaks, and reduce costs due to lower energy usage and taxes associated with energy and carbon emissions.

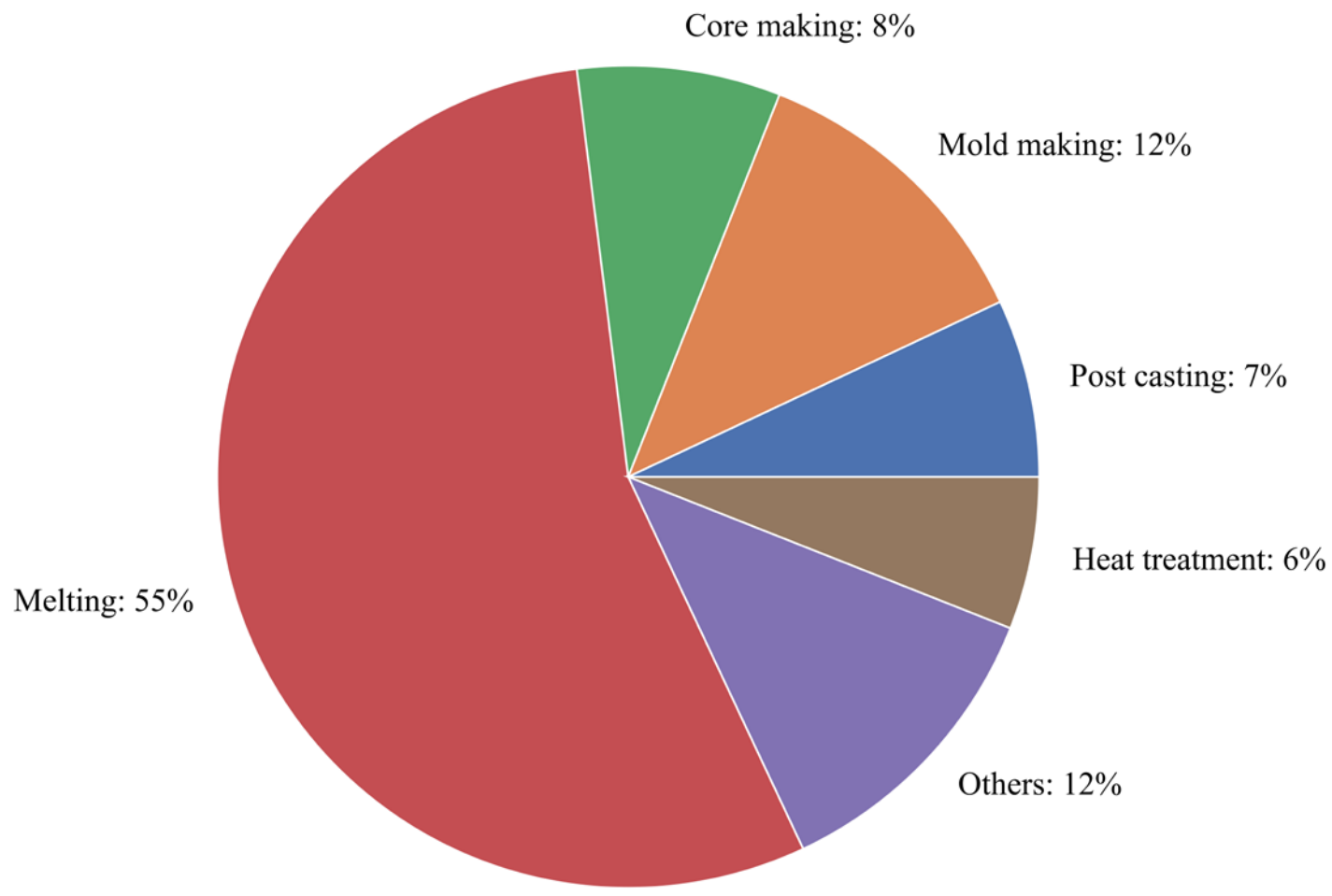

**Fig. 1.** Foundry industry energy consumption distribution based on [10]

Within the foundry industry, the operation is primarily based on the tacit knowledge of furnace operators [12]. To promote efficient operation, there is a need to identify the melting patterns that can be considered best practices. The diversity of melting patterns makes it difficult to categorize and analyze them effectively. Additionally, the lack of a systematic approach for evaluating and comparing the performance of different melting patterns hinders the identification of best practices. These factors highlight the need for a comprehensive, data-driven methodology to address these challenges. While previous studies have explored various aspects of energy efficiency and optimization in industrial processes, the specific problem of identifying best practice melting patterns in induction furnaces has not been adequately addressed in the literature.

Therefore, this paper aims to present a systematic method for identifying best-practice melting patterns in induction furnaces to enable energy-efficient operation. The method includes applying time-series K-means clustering to categorize melting patterns into clusters, calculating performance parameters for each cluster to assess their efficiency and environmental impact, employing multi-criteria decision-making methods to determine the best practice melting pattern cluster, and lastly evaluating the potential cost savings and energy efficiency improvements resulting from implementing the identified best practice pattern. The method is demonstrated in a case study of a Danish foundry.

The rest of this paper is structured as follows, initially, relevant background and literature is presented surrounding the current development of realizing energy efficiency in the foundry industry. Subsequently, the methodologies employed in this paper are outlined, providing an overview of the approach used to identify best practice melting patterns in induction furnaces. Afterward, the results are presented, showcasing the application of time series K-means clustering to categorize melting patterns and determine the ideal number of clusters using the elbow method. Next, relevant performance parameters are established for the clusters, along with the introduction of multi-criteria decision-making (MCDM) methods employed to evaluate and compare the performance of different clusters. Lastly, the potential cost savings and energy efficiency improvements resulting from implementing the best practice pattern are explored before discussing the results and concluding upon the findings with suggestions for further research.

## 2 Background

The fundamental process flow observed within a foundry production process can be shown in Fig. 2. As shown in Fig. 2, several production steps are involved in producing the final casting workpieces.

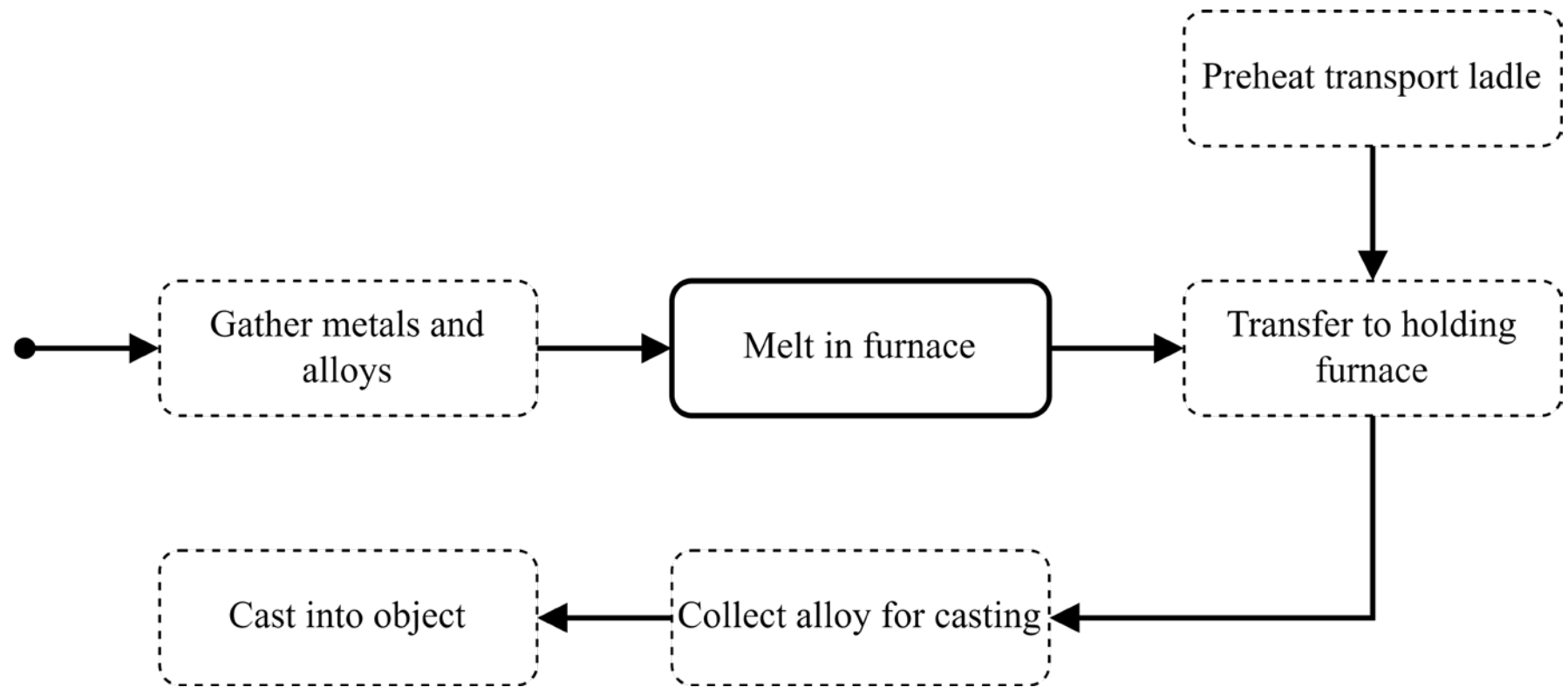

**Fig. 2.** Overview of the foundry production process

The foundry process shown in Fig. 2, is initiated with collecting and sorting melting material, primarily iron, and alloys, in different qualities, including the addition of scrap metal collected from the scrap metal pits. Subsequently, the iron is added to the induction furnace, where it is heated to the specified temperature adjusting for ferromagnetic losses. Once the melt has achieved the specified temperature, it is transferred to a pre-heated transfer ladle. In the transfer ladle, doping may be performed to achieve specific alloy capabilities adhering to the melt currently in the holding furnace. The melt is then transferred to a holding furnace, where it can be stored for a set period. The transfer to the holding furnace marks the completion of the melting part of the foundry process. Subsequently, the primary forming stage commences.

Once the primary forming starts, alloy of the required quality is collected from the holding furnace in a pre-heated transfer ladle. At this stage, doping can be performed again to meet customer-specific demands for alloy capabilities. Once the correct alloy has been achieved, it is transferred to one of the molding machines that perform the primary forming of the melt using sand-molded casting. The casting is subsequently transferred to a cooling line which marks the end of the primary forming of the casting.

## 2.1 Energy Efficiency in Foundry Production

Foundry production is a highly energy-intensive process that consumes large amounts of electricity, natural gas, and other fuels to melt and shape metal alloys. Over the past few decades, there has been growing interest in improving the energy efficiency of foundry production to reduce energy consumption, greenhouse gas emissions, and costs.

Several studies have focused on identifying the primary energy consumption sources in foundry production, such as melting and casting operations, heat treatment, and material handling. For example, a study found that melting and casting processes accounted for the majority of energy consumption in a steel foundry [10, 13]. However, it was found that heat treatment and melting accounted for the majority of energy consumption, approximately 60 %, in an aluminum foundry [14]. These studies have also identified various energy-saving opportunities, such as using more energy-efficient equipment, improving process control, and recycling waste heat.

Various strategies have been proposed to improve energy efficiency in foundries. One approach is to use more efficient equipment and technologies, such as induction melting and high-pressure die casting. Another strategy is to implement energy management systems and conduct energy audits to identify and address areas of inefficiency [14, 15].

Several studies have also applied advanced modeling and simulation techniques to optimize foundry production processes for energy efficiency. For instance, the use of computational fluid dynamics (CFD) and finite element analysis (FEA) have been used in several studies to optimize the design of melting and casting operations and to study the heat transfer processes within the furnace and mold, allowing for the identification of potential energy savings [16, 17]. Optimization techniques such as mathematical programming, artificial intelligence, and machine learning have also been reported to reduce energy consumption in the foundry process [18-21].

Foundry production process modeling and simulation have also been investigated to improve cost and energy efficiency. E.g., early foundry simulation efforts examined the impact of scheduled jobs in a foundry setting where each job should follow specific criteria relating to the number of castings needed and the weight of each casting [22]. Several studies have applied simulation in various aspects of the production process especially emphasizing molding, casting, and core shooting [12, 23-25].

Best-practice operations have been shown to enable substantial energy savings [26]. However, the concept of best practice operation in the context of foundries has not been sufficiently investigated in previous literature. Operational practices have been shown to impact the industry's performance and potential energy efficiency [27]. Previous studies have suggested that induction furnaces may experience up to 25 - 30 % losses due to unfavorable operation [28], considering that the majority of energy consumption in the foundry is consumed in the melting process; this presents a significant gap in the literature for promoting energy-efficient operation in the foundry.

This paper proposes clustering to identify and group melting patterns in the induction furnace to distinguish operational practices. In the literature, clustering has been used to identify energy consumption patterns in production processes and group simi-lar processes together based on their energy consumption profiles, e.g., [29]. This information has been used to identify opportunities for energy savings by optimizing the energy consumption of the production processes. Clustering has also seen use for benchmarking in energy regulation [30]. Clustering has also been used to group similar processes together based on their flexibility and responsiveness; to identify opportunities for process optimization and improvement. For instance, clustering can identify processes that can switch between energy sources or operate at different production rates, enabling them to respond to energy availability or demand changes, e.g., [31]. Lastly, clustering has seen use in identifying the optimal configuration of production processes to maximize energy efficiency and flexibility. Clustering can identify the optimal configuration of production processes by analyzing the relationships between different variables, such as energy consumption, production rates, and operating conditions, e.g., [31].

MCDM has been proposed for establishing best practices in various domains, such as quality management and selection of technologies [32, 33]. However, it has not seen in-depth investigation in establishing industry best practices. However, MCDM has seen usage for establishing energy efficiency practices. Previous studies have examined the use of MCDM in the automotive industry to improve energy efficiency in automotive engineering and service center selection [34, 35]. Furthermore, in the beverage industry, MCDM has seen use for choosing energy improvement measures [36]. MCDM has also seen use in selecting locations for energy storage systems in combination with using K-means++, and for mapping industries for participation in electricity markets [37, 38]. In summary, it is evident that there is a need to establish best practice operations for furnace operation in foundries. Best practices have been established both through clustering and MCDM in various domains; however, there is little literature on combining the methods for providing industries with the identification of best practice operations. Utilizing a data-driven clustering approach ensures that the best practice adheres to the constraints of the process as the clusters build on historical data, while the MCDM allows the facility to prioritize the subjective weighting of what performance parameters indicate best practice.

## 3 Methodology

An overview of the methodology used in this paper can be seen in Fig. 3. To enable the clustering of operational practices and identification of the best practice cluster, the relevant clustering and MCDM algorithms are selected.

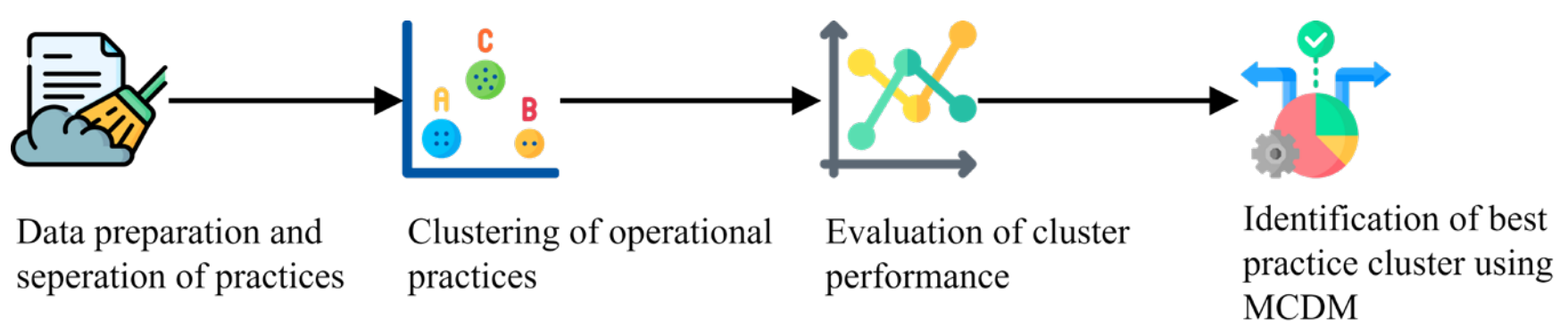

**Fig. 3.** Methodology for identifying best-practice melting patterns.

### 3.1 Clustering Algorithm Selection

As production processes often collect time-series data and experience variance in the processing time, the data requires specialized clustering algorithms that can handle the temporal nature of the data and capture the patterns and relationships within the time series. Therefore, time-series clustering is proposed, as time-series clustering is an unsupervised learning technique used to identify patterns and group similar data points based on their temporal behavior. Previous studies have compared various clustering algorithms; it was found that the K-means algorithm provided better performance while also being computationally efficient [39]. Time-series K-means clustering is a modified version of K-means clustering that considers the temporal nature of the data. This method involves representing each time series as a sequence of vectors and clustering these vectors using K-means clustering; the vectors may be established using dynamic time warping or Euclidian distance [40]. The resulting clusters represent similar patterns in the time-series data.

Compared to other clustering algorithms, time-series K-means clustering has several advantages. Previous comparisons have established that the performance of the K-means clustering algorithm outweighs other clustering algorithms, and furthermore, the implementation of K-means is faster compared to other algorithms [40-42]. This is further supported by the ability to scale with large datasets and guaranteed convergence [43]. Other clustering algorithms, such as Kernel K-means, and Kshape, have also been used for time-series clustering. However, these algorithms have limitations when dealing with the computation of cluster centroid and datasets of various lengths due to relying on cross-correlation of the cluster time series [40].

The elbow method is commonly used to determine the optimal K-means clustering cluster size. The elbow method involves plotting the within-cluster sum of squares against the number of clusters and selecting the number of clusters at the "elbow" of the plot, where the rate of decrease in the within-cluster sum of squares slows down [44]. The within-cluster sum of squares measures the sum of the squared distances between each data point and its assigned cluster center. The idea behind the elbow method is to select the number of clusters that significantly decreases the within-cluster sum of squares while minimizing the number of clusters [44]. Increasing the number of clusters too much can lead to overfitting and loss of generalizability, while too few clusters may not capture all the underlying patterns in the data. Time series K-means clustering was utilized in this paper to identify clusters of similar profiles relating to the operation. By identifying clusters of operational patterns the performance of the various practices could be evaluated to identify efficient and inefficient operations.

### 3.2 Multi-criteria Decision Making

The performance of a specific operation may be evaluated across several parameters, and the goals and preferences of various foundries may vary. Therefore, there may not be a single solution that fits all foundries; therefore, MCDM algorithms are imple-mented to incorporate the goals and preferences of the foundry when evaluating a specific operational practice.

MCDM is a method used to select the best option from a set of alternatives based on multiple criteria or objectives [45]. There are several MCDM techniques available, such as Analytical Hierarchy Process (AHP), the Technique for Order Preference by Similarity to Ideal Solution (TOPSIS), and VlseKriterijumska Optimizacija I Kompromisno Resenje (VIKOR). The MCDM process involves identifying the criteria, weighing them according to their importance, and evaluating the alternatives against the criteria [45]. In the context of energy usage in industrial production processes, MCDM can be applied to support decision-making in various scenarios, such as production planning, resource allocation, and energy performance evaluation, e.g. [46]. The MCDM approach has several advantages, such as the ability to handle multiple criteria, the flexibility to incorporate subjective preferences, and the ability to evaluate alternatives comprehensively. However, MCDM also has some limitations, such as difficulty determining the appropriate weights for the criteria and the subjective nature of the decision-making process [47].

This paper has implemented a series of MCDM algorithms based on the work conducted in [48] and [49]. The MCDM algorithms utilized in this research are Simple Additive Weighting (SAW), Multiplicative Exponential Weighting (MEW), TOPSIS, Modified Technique for Order Preference by Similarity to Ideal Solution (mTOPSIS), and VIKOR. By utilizing multiple MCDM algorithms, the robustness of the final decision can be increased, and any differences in ranking can be examined [50].

## 4 Case Study

A large Danish foundry provides a case study for the application of the identification of best-practice operations. The Danish foundry is the largest in Northern Europe and produces 45,000 tonnes of casting products each year, exported to 25 different countries. The foundry has committed to reducing its greenhouse gas emissions and has actively implemented circular economy and sustainability as active goals in its business strategy. As a part of the foundry's sustainability efforts, the energy consumption of their production was mapped. The mapping revealed that 78.5 % of their annual energy consumption is electricity, predominantly consumed by the melting and holding furnaces. As part of the energy mapping, it was shown that 0.5 tonnes of carbon dioxide emissions were emitted per tonne of produced casting.

The case study is ideal for identifying best-practice melting patterns due to their commitment to improving sustainable foundry practices and being a state-of-the-art facility utilizing induction furnaces with processes monitored through existing sensors. The production observed in the foundry case study follows the steps shown in Fig. 2. The facility includes an in-house factory where machining and surface treatment of casting workpieces can be undergone. In this study, only the foundry process is considered as this is the mandatory part of the production flow. This study focuses on the melting operation in the furnace, and the adjacent steps shown with dashed lines are not included.

## 5 Results

Based on the case study, data for one of the induction furnaces could be obtained. The period range of the obtained data was from the 11th of May 2022 to the 30th of May 2022. An overview of the parameters and data completeness can be seen in Fig. 4.

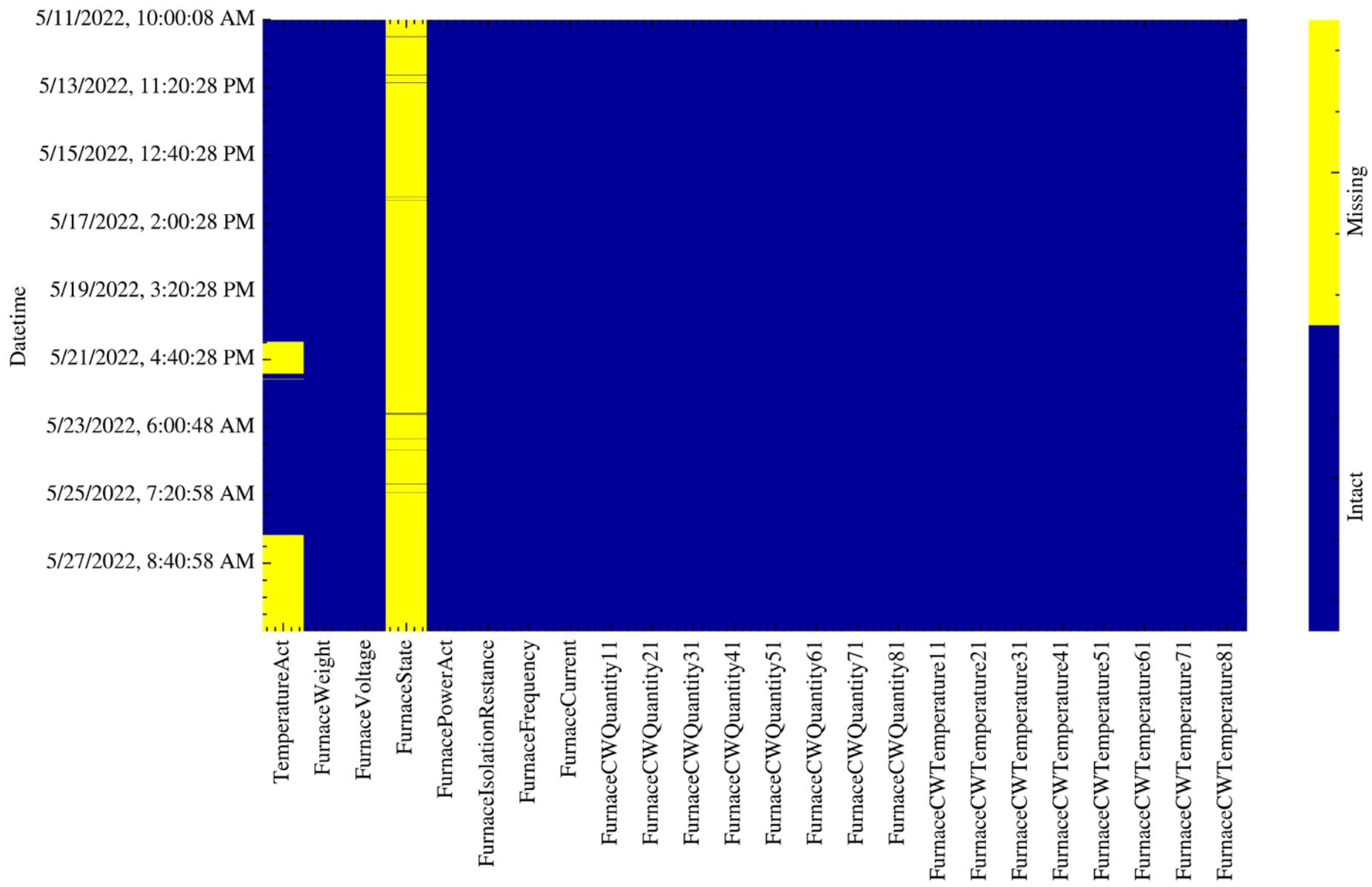

**Fig. 4.** Induction furnace parameters and data completeness.

Before initiating the cluster identification, the data was cleaned and prepared. The cleaning involved removing the furnace state parameter and removing the rows of data with missing temperature measurements. The parameters collected in Fig. 4 include the melt temperature, melt weight, furnace voltage, furnace state, furnace power consumption, furnace isolation resistance, furnace frequency, furnace current, and multiple measuring points for the cooling water temperature and flow.

Using temperature to identify melting patterns in induction furnaces is driven by its significance in melting. Temperature is a fundamental parameter that directly impacts the physical and chemical transformations occurring within the furnace and provides a comprehensive representation of the thermal behavior of the furnace during the melting cycle. Examining temperature data over time enables the identification of distinct patterns in the melting process. These patterns can be associated with different operational conditions, such as the charging of materials, power input, or changes in furnace load. In contrast, parameters like power consumption may not capture the nuances of the melting process as effectively as temperature. Therefore, the temperature profiles were investigated and can be seen in Fig. 5.

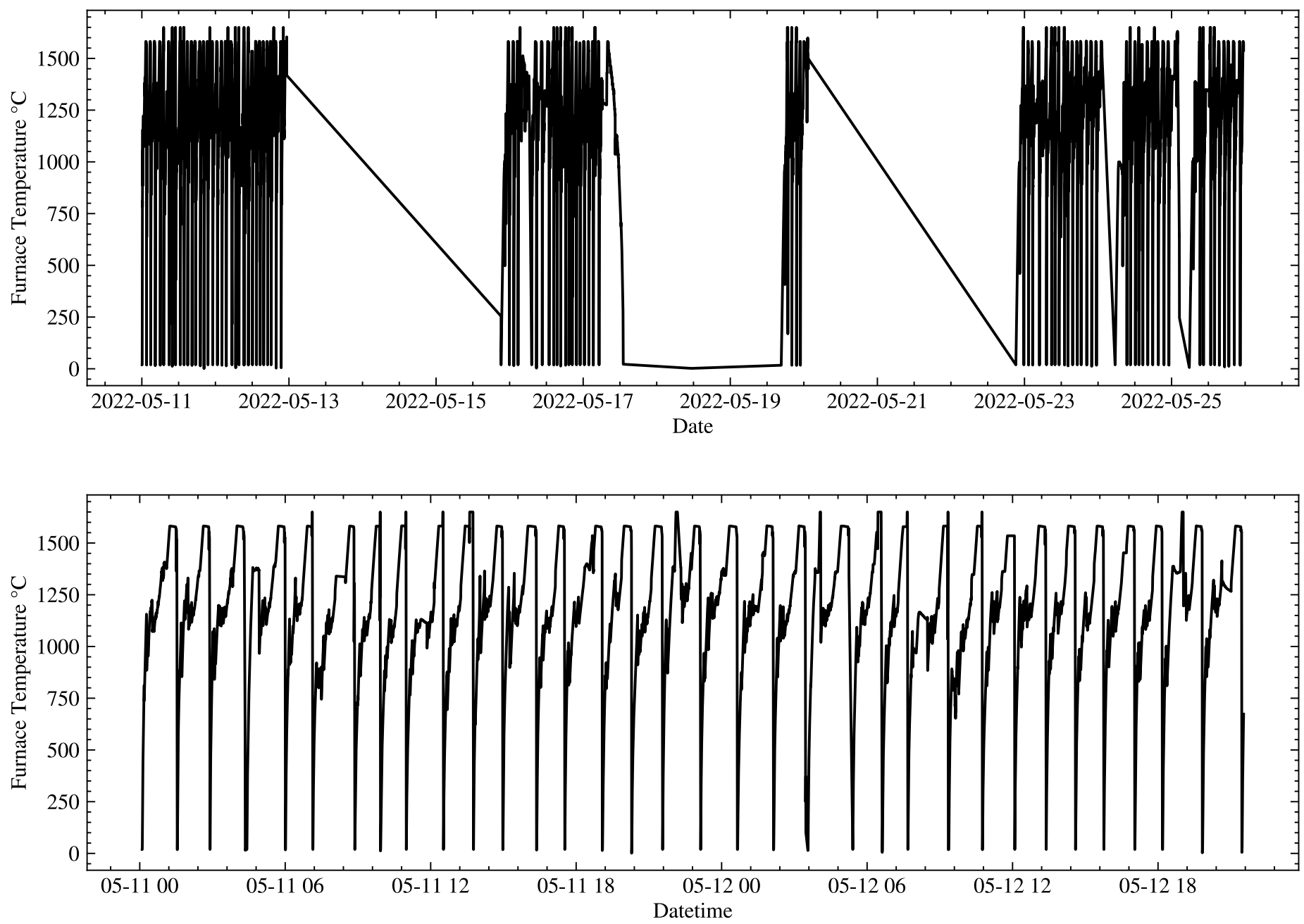

**Fig. 5.** Time-series overview of the melt temperature from the furnace computer for the whole period and for a segment of the period

As the obtained data is in a time-series format, an algorithm is necessary to separate the individual melts. The algorithm used to separate the time series data in this study can be seen in Fig. 6.

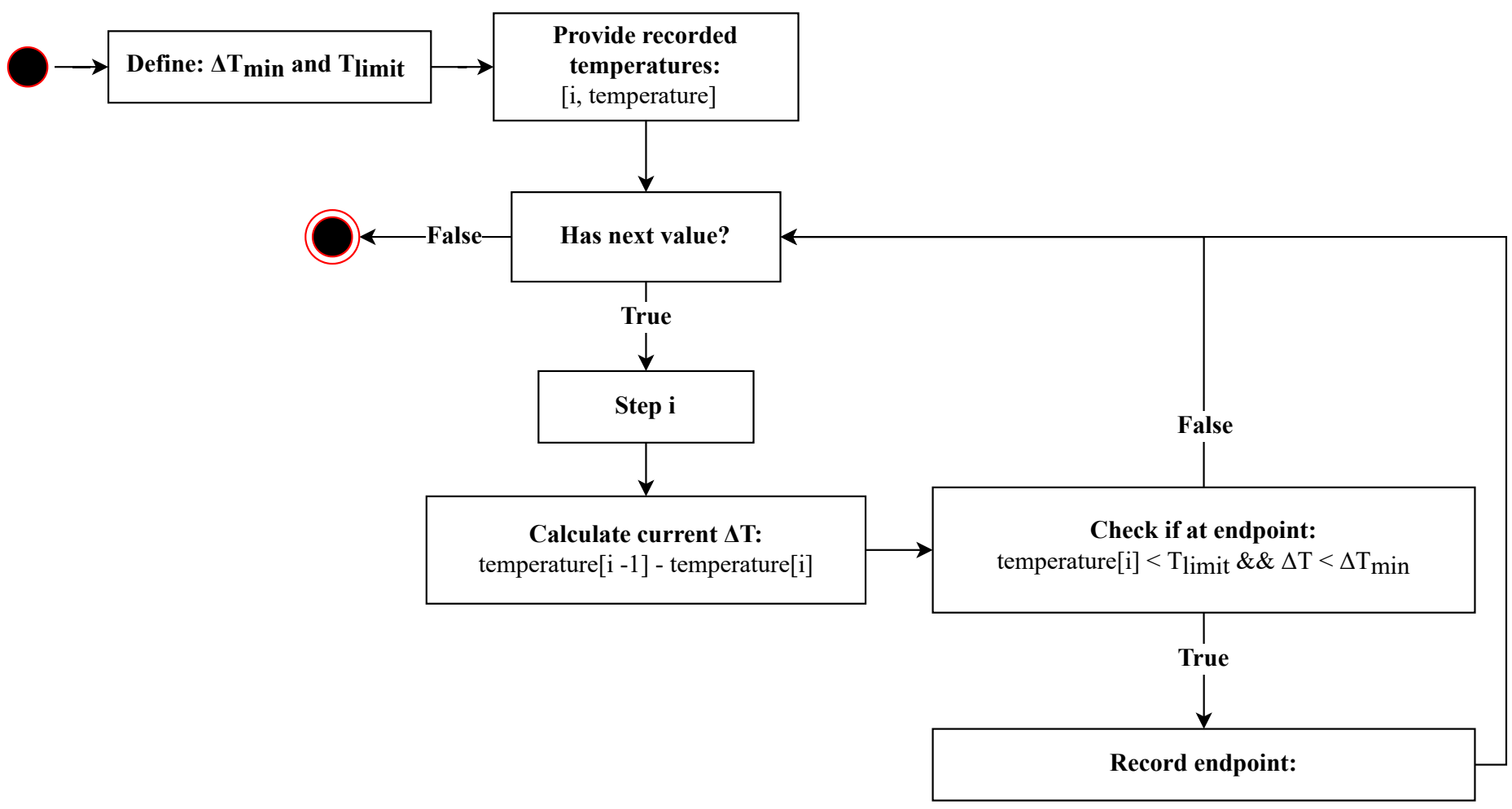

**Fig. 6.** Melting pattern identification algorithm.

The algorithm seeks to identify the melting endpoints from which the starting points of the subsequent melt can be inferred. Using the index, the change in temperature from one time to the next can be calculated. As shown in Fig. 5, a significant decrease in temperature is observed at the end of a melt by tuning the minimum temperature, and the minimum change in temperature of the melts can be identified. The limits are imposed as changes in temperature can happen due to adding new material to the melt, and the change should hence occur under a set temperature limit, indicating that the melt has been completed. Ninety-three individual melts were identified for the period.

### 5.1 Time-series K-means Clustering of Melting Profiles

Timeseries K-means clustering was performed to identify similar melting profiles. As seen in Fig. 7, the number of clusters was determined based on the elbow method with a comparison of the inertia, distortion, and silhouette scores for finding the optimal K.

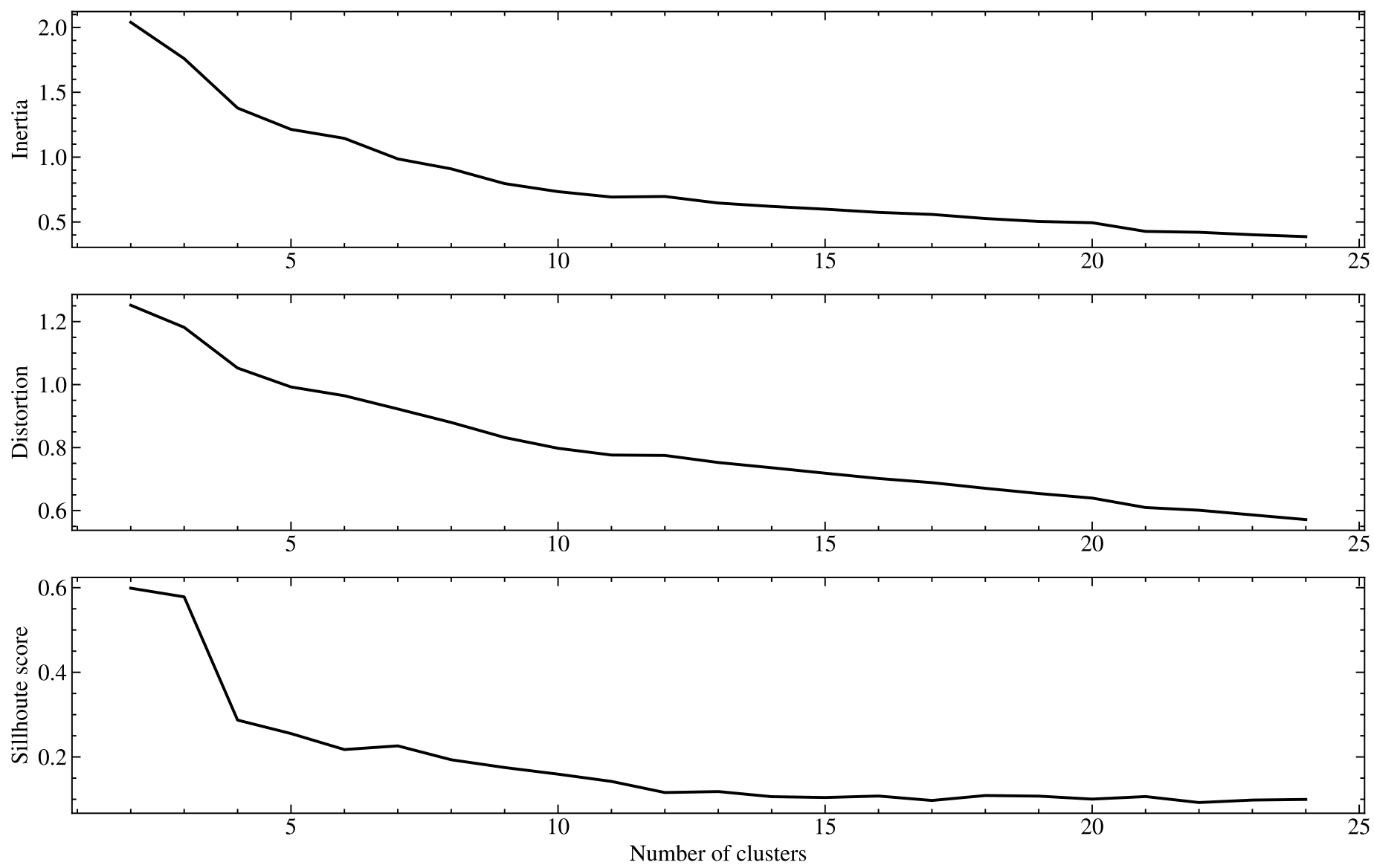

**Fig. 7.** Inertia, Distortion, and Silhouette score in response to the number of clusters

As shown in Fig. 7, the inertia and distortion do not yield a distinct elbow. Examining the silhouette score shows an elbow at K = 12. Subsequently, the time series K-means clustering was performed using the number of clusters determined from the silhouette score. The result of the clustering can be seen in Fig. 8.

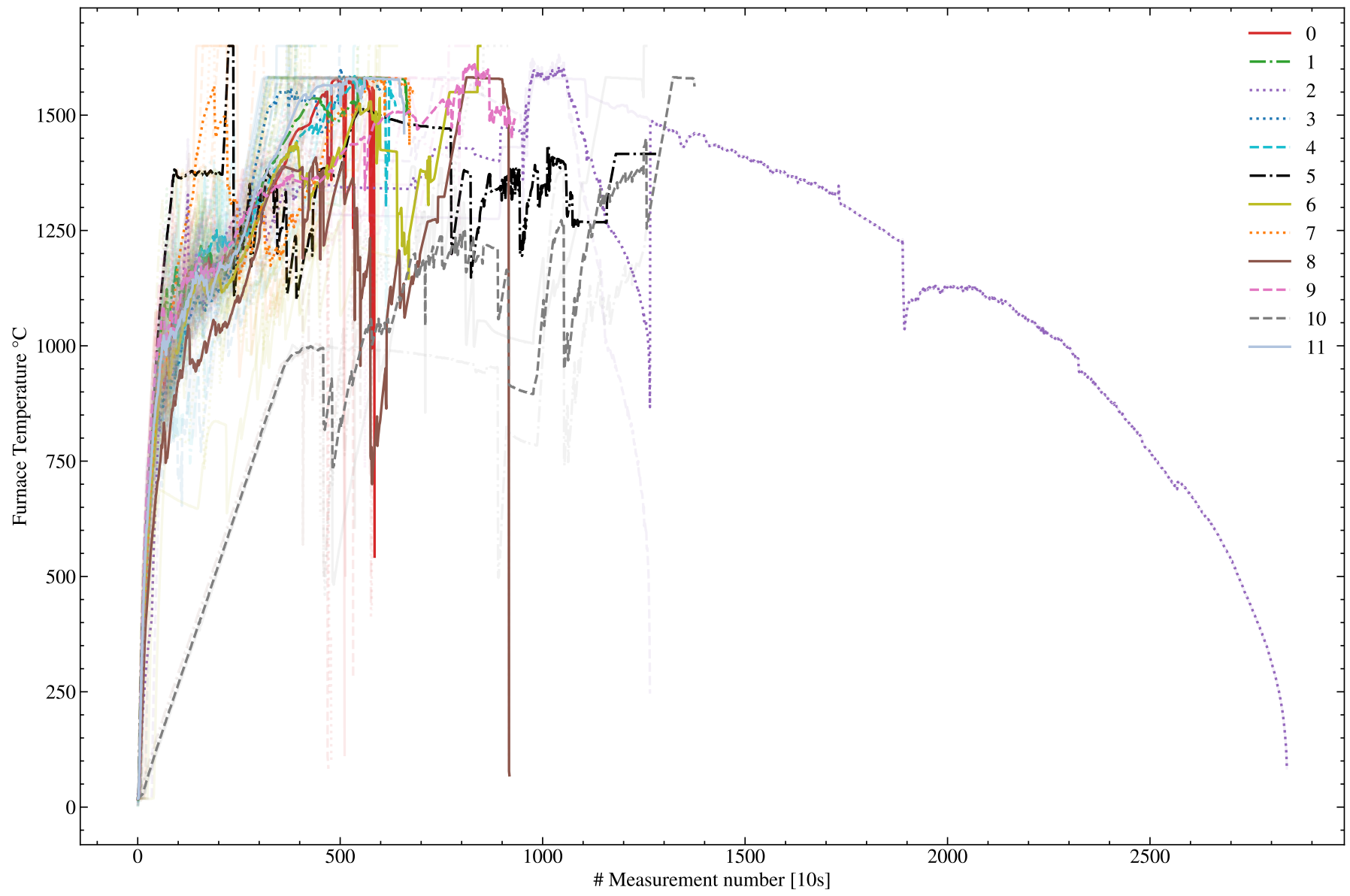

**Fig. 8.** Clustering of melting profiles showcasing the average of each cluster.

As seen in Fig. 8, 12 clusters are identified, with each cluster plot line showing the mean temperature of all current cluster members at a given time step. Each cluster consists of several melting profiles; the distribution of melting profiles in each can be seen in Fig. 9.

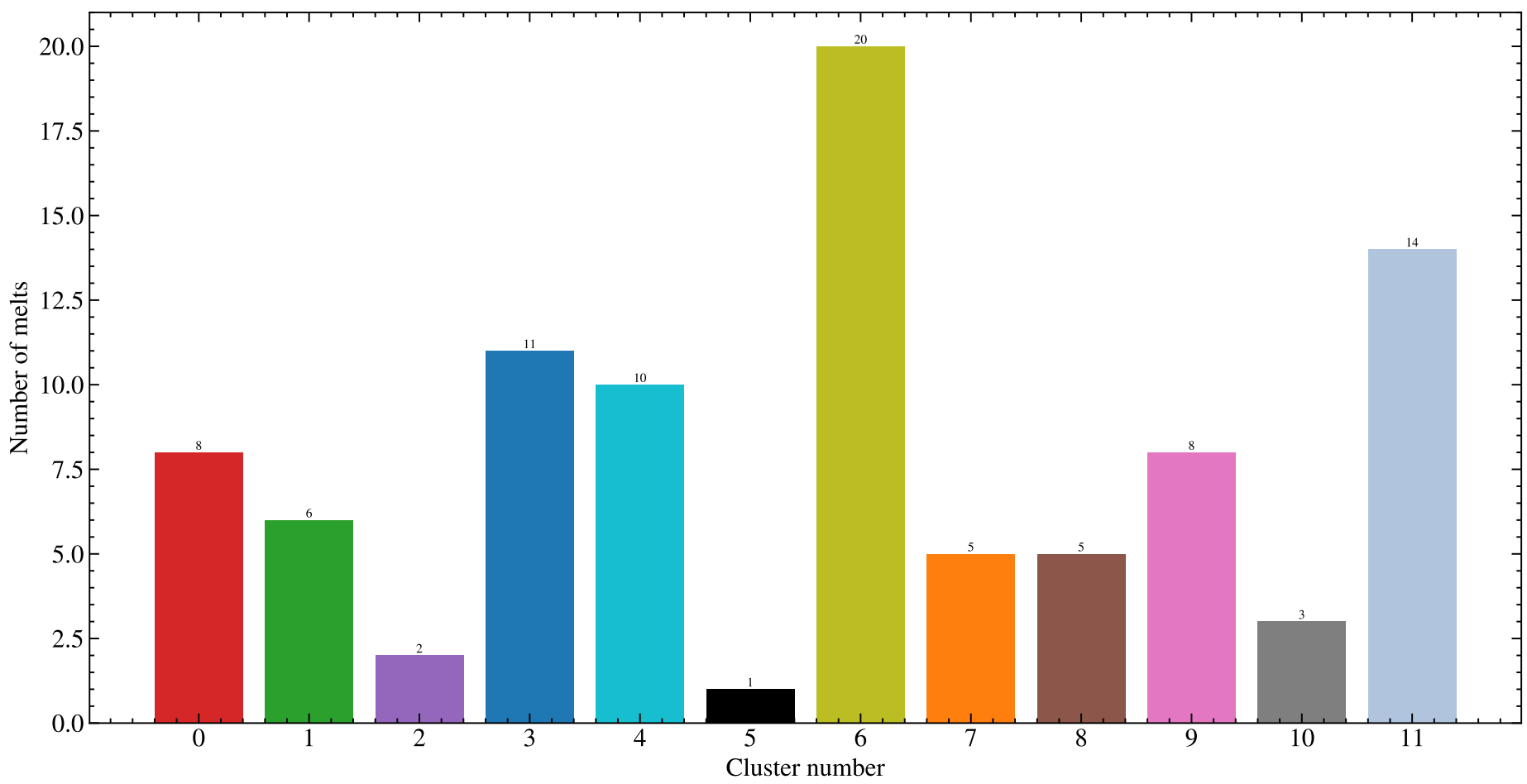

**Fig. 9.** Distribution of the number of melts within each cluster

As seen in Fig. 9, the melts were distributed in 12 clusters, with cluster no.6 containing the highest number of melts. The clusters containing the highest number of melts (e.g., no. 3, 4, 6, and 11) represent the common operational practice at the furnace. Meanwhile, other clusters, e.g., no. 2, 5, and 10 represent uncommon operational practices. According to the furnace operators' explanation, the few melts in cluster no.2 were due to changing the furnace's refractory lining, i.e., scheduled furnace maintenance. With further collaboration with the furnace operators, specific operational practices were identified within the identified clusters, aiding in explaining and understanding the observed cluster melting pattern.

### 5.2 Multi-criteria Decision Making for Best Practice Melting Profile Identification

As established in the previous chapter, the clusters contain similar melting profiles. Each melting profile has an associated performance in terms of melting time, energy consumption, etc. Therefore, to sufficiently assess which cluster should be considered best practice, MCDM is used. MCDM allows assessing an array of options with associated parameters, where each parameter may be weighted differently. Using MCDM hence enables the foundry to emphasize the performance metrics that are important to them. The cluster-specific performance metrics can be seen in Table 1.

**Table 1.** Cluster-specific performance parameters

| Cluster | Average production time [s] | Average electricity consumption [kWh] | Average energy-specific performance [kWh/tonne] | 2030 Carbon tax [DKK] |
|---|---|---|---|---|
| 0 | 5346.25 | 5186.90 | 554.83 | 509.44 |
| 1 | 5251.67 | 5111.82 | 552.55 | 552.57 |
| 2 | 20515.00 | 5364.75 | 573.80 | 643.60 |
| 3 | 4332.73 | 4858.24 | 522.00 | 338.01 |
| 4 | 5221.00 | 4980.30 | 551.26 | 496.16 |
| 5 | 12870.00 | 6368.12 | 664.31 | 803.86 |
| 6 | 4892.00 | 4880.48 | 541.03 | 396.21 |
| 7 | 6216.00 | 5236.36 | 554.28 | 493.01 |
| 8 | 5886.00 | 5028.03 | 550.13 | 368.42 |
| 9 | 7370.00 | 5027.46 | 563.93 | 404.34 |
| 10 | 11820.00 | 5260.52 | 605.99 | 410.19 |
| 11 | 5218.57 | 5065.82 | 545.92 | 421.69 |

Several MCDM methods were used to compare the performance of the individual clusters. Table 2 shows the ranking of each of the MCDM methods. Equal weighting was assumed between each performance parameter.

**Table 2.** Cluster-specific MCDM ratings

| Cluster | SAW | MEW | TOPSIS | mTOPSIS | VIKOR |
|---|---|---|---|---|---|
| 0 | 0.25783 | 0.25160 | 0.82853 | 0.82853 | 0.30858 |
| 1 | 0.26195 | 0.25443 | 0.79901 | 0.79901 | 0.35786 |
| 2 | 0.40147 | 0.37967 | 0.17793 | 0.17793 | 0.83386 |
| 3 | 0.21649 | 0.20874 | 1.00000 | 1.00000 | 0.00000 |
| 4 | 0.25162 | 0.24555 | 0.84409 | 0.84409 | 0.26624 |
| 5 | 0.38969 | 0.38676 | 0.37868 | 0.37868 | 1.00000 |
| 6 | 0.23199 | 0.22617 | 0.93427 | 0.93427 | 0.11052 |
| 7 | 0.26293 | 0.25968 | 0.81689 | 0.81689 | 0.29767 |
| 8 | 0.23903 | 0.23532 | 0.90686 | 0.90686 | 0.16566 |
| 9 | 0.25763 | 0.25636 | 0.82027 | 0.82027 | 0.25173 |
| 10 | 0.30212 | 0.29815 | 0.59682 | 0.59682 | 0.50407 |
| 11 | 0.24141 | 0.23617 | 0.90259 | 0.90259 | 0.16635 |

As can be seen from the table above, cluster three (grey) shows the overall best performance as it scores best across all MCDM methods. Hence, cluster three is selected as the cluster representing the best-practice operation.

### 5.3 Energy Efficiency Potential for Best-practice Operation

The identified best-practice cluster performance was assumed across all melts in the investigated period to assess the impact of using best-practice operations, assuming 100 % adherence to the best-practice operation. I.e. the average performance of the melts in the best practice cluster were assumed for all other clusters in the dataset. Using the recorded start times of the melts in the other clusters the corresponding electricity cost and $CO_2$ emissions could be calculated subject to the best practice performance. The foundry is assumed to follow the spot market electricity price and $CO_2$ emissions. The implication of utilizing best practice operations can be seen in Table 3.

**Table 3.** Comparison of current operation and implementation of best practice operation

| Operating Mode | Electricity Cost [DKK] | 2030 Carbon Cost [DKK] | CO2 Emissions [kg] | Total Electricity Cost [DKK] |
|---|---|---|---|---|
| Current practice | 602913.00 | 40761.88 | 54349.18 | 643674.89 |
| Best practice | 551041.56 | 37321.17 | 49761.56 | 588362.73 |
| Percentage change | 8.60 % | 8.44 % | 8.44 % | 8.59 % |

## 6 Discussion

To identify best-practice melting patterns in induction furnaces, the following four steps have been conducted:

1. Application of the time series K-means clustering to categorize melting patterns into clusters
2. Calculation of the performance parameters for each cluster to assess their efficiency and environmental impact
3. Deployment of the multi-criteria decision-making methods to determine the best practice melting pattern cluster,
4. Evaluation of the potential cost savings and energy efficiency improvements resulting from implementing the identified best practice pattern.

In accordance with the presented method, the obtained data was prepared and separated into individual melts. The application of time-series K-means clustering enabled the identification and categorization of melting patterns in the induction furnace, leading to the determination of an optimal number of clusters using the elbow method, as seen in Fig. 7. Examining the distribution of clusters from Fig. 9 and the cluster patterns in Fig. 8 furthermore enabled identification of specific operations within the furnace, e.g., change of refractory lining. The cluster distribution furthermore showed that the melts adhering to the best practice operation only occur around 12 % of the time. As seen in Table 1, the performance parameters calculated for each cluster, such as melting time, energy-specific performance (kWh/tonne), and carbon cost, provided insights into different patterns' energy efficiency and cost-effectiveness. Using MCDM methods, SAW, MEW, TOPSIS, mTOPSIS, and VIKOR, facilitated comparing and selecting a cluster representing the best-practice performance seen in Table 2. The foundry could achieve cost savings by implementing the identified best practice melting pattern, with an estimated reduction of approximately 8.6% in electricity costs.

## 7 Conclusion

This paper has addressed the challenge of identifying best practice melting patterns in induction furnaces through a data-driven methodology utilizing time-series K-means clustering and multi-criteria decision-making (MCDM) methods. The cluster representing the best practice performance could be identified by categorizing melting patterns into clusters and evaluating their performance based on various parameters such as melting time, energy-specific performance, and carbon cost.

The implications of the findings are significant for the foundry industry, as implementing the identified best practice melting pattern can lead to substantial cost savings and improved energy efficiency. The estimated electricity cost savings of approximately 8.6% demonstrate the tangible benefits of improving melting practices.

Building on the findings presented in this paper, there is potential for exploring further research. The findings in this paper build on the investigation of a single furnace within the foundry, which could be extended to include other furnaces. Furthermore, limited information was available on the alloy composition within the furnace; for future research, it would be beneficial to include the alloy composition to investigate the impact of alloy composition on best practice operation.

From the methodology perspective, investigating the scalability and applicability of the methodology for the identification of best practice melting patterns across different foundries and regions would aid in verifying the robustness of the methodology. Furthermore, the applicability of the methodology outside of the foundry domain could be investigated to examine the potential for the identification of best-practice operations in other domains. Lastly, the implementation of the best practice operation is subject to underlying uncertainties present in production processes; therefore, it would be beneficial to develop a simulation model that can accurately represent the process flow and evaluate any unforeseen consequences due to the best practice operation to provide a more nuanced understanding of the best practice implications.

**Acknowledgment**

This paper is part of the "IEA-IETS Annex XVIII: Digitization, artificial intelligence and related technologies for energy efficiency and reduction of greenhouse gas emissions in industry" (Case no. 134-21010), and "Data-driven best-practice for energy-efficient operation of industrial processes - A system integration approach to reduce the CO2 emissions of industrial processes" (Case no.64020-2108) funded by the Danish Energy Technology Development and Demonstration (EUPD) program, Denmark.